\documentclass{article}

\usepackage{arxiv}

\usepackage[utf8]{inputenc} % allow utf-8 input
\usepackage[T1]{fontenc}    % use 8-bit T1 fonts
\usepackage{hyperref}       % hyperlinks
\usepackage{url}            % simple URL typesetting
\usepackage{booktabs}       % professional-quality tables
\usepackage{amsfonts}       % blackboard math symbols
\usepackage{amsmath}
\usepackage{nicefrac}       % compact symbols for 1/2, etc.
\usepackage{microtype}      % microtypography
\usepackage{cleveref}       % smart cross-referencing
\usepackage{lipsum}         % Can be removed after putting your text content
\usepackage{graphicx}
\usepackage{natbib}
\usepackage{doi}

\usepackage{subcaption}

\usepackage{float}

\usepackage{caption} 
\usepackage{booktabs}
\usepackage{multirow}
\usepackage{array}
\newcolumntype{L}[1]{>{\raggedright\let\newline\\\arraybackslash\hspace{0pt}}m{#1}}
\newcolumntype{C}[1]{>{\centering\let\newline\\\arraybackslash\hspace{0pt}}m{#1}}

\usepackage{bm}
\usepackage{siunitx}

\title{Explainability through uncertainty: Trustworthy decision-making with neural networks}

\date{}

\author{ \href{https://orcid.org/0000-0001-9107-5646}{\includegraphics[scale=0.06]{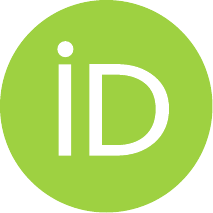}\hspace{1mm}Arthur Thuy}\thanks{Corresponding author} \\
	Ghent University\\
	CVAMO Core Lab, Flanders Make\\
	\texttt{arthur.thuy@ugent.be} \\
	\And
	\href{https://orcid.org/0000-0001-9901-8507}{\includegraphics[scale=0.06]{orcid.pdf}\hspace{1mm}Dries F.~Benoit} \\
	Ghent University\\
	CVAMO Core Lab, Flanders Make\\
	\texttt{dries.benoit@ugent.be} \\
}

\renewcommand{\headeright}{Accepted Manuscript version of an article published in the European Journal of Operational Research}
\renewcommand{\undertitle}{Accepted Manuscript version of an article published in the European Journal of Operational Research (\href{https://doi.org/10.1016/j.ejor.2023.09.009}{\texttt{10.1016/j.ejor.2023.09.009}})}
\renewcommand{\shorttitle}{}

\hypersetup{
	pdftitle={Explainability through uncertainty: Trustworthy decision-making with neural networks},
	pdfsubject={cs.LG, stat.ML},
	pdfauthor={Arthur Thuy, Dries F.~Benoit},
	pdfkeywords={Decision support systems, Explainable artificial intelligence, Monte Carlo Dropout, Deep Ensembles, Distribution shift},
}

\begin{document}
	\maketitle
	
	\begin{abstract}
		Uncertainty is a key feature of any machine learning model and is particularly important in neural networks, which tend to be overconfident.
		This overconfidence is worrying under distribution shifts, where the model performance silently degrades as the data distribution diverges from the training data distribution.
		Uncertainty estimation offers a solution to overconfident models, communicating when the output should (not) be trusted.
		Although methods for uncertainty estimation have been developed, they have not been explicitly linked to the field of explainable artificial intelligence (XAI). 
		Furthermore, literature in operations research ignores the actionability component of uncertainty estimation and does not consider distribution shifts.
		This work proposes a general uncertainty framework, with contributions being threefold: (i) uncertainty estimation in ML models is positioned as an XAI technique, giving local and model-specific explanations; (ii) classification with rejection is used to reduce misclassifications by bringing a human expert in the loop for uncertain observations; (iii) the framework is applied to a case study on neural networks in educational data mining subject to distribution shifts. 
		Uncertainty as XAI improves the model's trustworthiness in downstream decision-making tasks, giving rise to more actionable and robust machine learning systems in operations research.
	\end{abstract}

	% keywords can be removed
	\keywords{Decision support systems \and Explainable artificial intelligence \and Monte Carlo Dropout \and Deep Ensembles \and Distribution shift}

% =====================================================================
% Main matter
% =====================================================================

\section{Introduction}
A representation of uncertainty is desirable and is a key feature of any machine learning (ML) model.
Uncertainty is particularly important in neural networks (NNs), which tend to be overconfident in their predictions \citep{guo2017calibration}.
That is, a NN classifier often predicts an incorrect label, despite giving a high predicted probability. 

This flaw is especially troubling in situations of distribution shift, where the data distribution during deployment diverges from the training data distribution \citep{murphy2022probabilistic}. 
Although the model performs well when first deployed, its performance degrades over time as the distribution shift increases without warning the decision maker. 
Distribution shifts happen all the time, either suddenly, gradually, or seasonally \citep{huyen2022designing}. 
For example, a demand prediction model is affected by a sudden change in the pricing policy of a competitor or when a new competitor enters the market.

The field of uncertainty estimation provides a solution to overconfident models by capturing the uncertainty in both the data and the model. 
As such, it communicates when a model's output should (not) be trusted \citep{ovadia2019can}. 
Building trust is also the cornerstone of the field of explainable artificial intelligence (XAI), which aims to explain the output of black-box models.
XAI techniques are commonly used in operations research (OR) to facilitate the human-computer interaction and thereby support decision-making systems \citep{cabitza2023quod}.

The related work on uncertainty estimation and XAI has three shortcomings: (i) uncertainty estimation is not explicitly formulated as an XAI technique following the local/global and model-specific/agnostic specification and there is no theoretical motivation on how uncertainty contributes to explainability; (ii) the available work in OR merely monitors the NN uncertainty estimates without acting upon it; (iii) there is a lack of OR applications that examine the influence of distribution shifts on NN uncertainty, as literature only employs benchmark datasets like MNIST.

A general uncertainty framework is proposed, with contributions being threefold:
\begin{enumerate}
	\item 
	The framework first positions uncertainty estimation in ML models as an \textit{XAI technique}, giving local and model-specific explanations.
	To support this, theoretical properties are discussed, arguing that uncertainty estimation fosters higher levels of \textit{trust}, \textit{actionability}, and \textit{robustness}.
	
	\item 
	The framework then uses \textit{classification with rejection} \citep{mena2021survey} to reduce misclassifications by bringing a human expert in the loop for uncertain observations.
	
	\item 
	The framework is applied to a case study on \textit{neural networks} in educational data mining, with distribution shifts occurring naturally when deploying the model to production.
\end{enumerate}

The remainder of the paper is organized as follows. 
Section \ref{sec:related_work} gives an overview of related work and identifies shortcomings.
Section \ref{sec:methodology} presents the general uncertainty framework and positions uncertainty estimation as an XAI technique.
Section \ref{sec:uncertainty_nn} discusses how uncertainty is quantified specifically in NN classifiers. 
In section \ref{sec:case_study}, the case study in educational data mining with NN uncertainty is presented; section \ref{sec:results} gives the results.
Finally, section \ref{sec:discussion} provides a discussion and section \ref{sec:conclusion} gives a conclusion.

\section{Related Work}
\label{sec:related_work}
This section discusses related work on XAI, uncertainty estimation, and NNs in the field of OR.
Furthermore, extant literature on uncertainty estimation as XAI is discussed.
Thereby, three main shortcomings in related work are identified.

\subsection{Explainable artificial intelligence in operations research}

ML models are widely used in OR to solve complex problems \citep{choi2018big}. 
However, extant literature often focuses on predictive performance which comes at the expense of model explainability. 
This lack of explainability leads to decision makers’ distrust and unwillingness to adopt analytics in decision support systems \citep{shin2021effects}.

The field of XAI refers to techniques that try to explain how a black-box ML model produces its outcomes.
Although still limited, XAI techniques are increasingly adopted in OR applications, e.g., in credit risk \citep{bastos2022explainable, sachan2020explainable}, marketing risk \citep{de2018new, van2020predicting}, supply chain management \citep{garvey2015analytical}, healthcare \citep{piri2017data}, and jurisprudence \citep{delen2021imprison}. 
As such, XAI bridges the gap to organizational decision-makers by providing understanding into a model’s predictions and generating actionable insights.

XAI techniques can be organized based on two main criteria \citep{adadi2018peeking}. 
It can be global, i.e., characterize the whole dataset (e.g., partial dependence plot), or local, i.e., explain individual predictions (e.g., counterfactual explanations). 
It can be model-specific, i.e., capable of explaining only a restricted class of models (e.g., random forest variable importance), or model-agnostic, i.e., applicable to any model (e.g., SHAP).

\subsection{Uncertainty and neural networks in operations research}

Neural networks are rapidly emerging in operations research (OR), with applications such as credit scoring, demand prediction, and outlier detection \citep{kraus2020deep, gunnarsson2021deep, verboven2021autoencoders, van2021using}. 
\citet{kraus2020deep} point to three key challenges that limit the relevance of deep learning in OR: (i) extensive hyperparameter tuning is required, (ii) lack of uncertainty estimation, and (iii) lack of accountability and explainability.

Uncertainty estimation for NNs has been investigated in different domains of OR: predictive maintenance \citep{kraus2019forecasting}, recommender systems \citep{nahta2021hybrid}, finance \citep{ghahtarani2021new}, stress-level prediction \citep{oh2021inductive}, transportation \citep{zhang2020bayesian, feng2022understanding}, predictive process monitoring \citep{weytjens2022learning}, and educational data mining \citep{yu2021exploring}.
In the available work, however, uncertainty estimates are merely monitored as an additional metric, not used in combination with a human expert such as in \textit{classification with rejection} (i.e., shortcoming 1). 
Ignoring the actionability of this human-machine combination leaves a large part of the added value on the table. 

Moreover, there is a lack of literature on the impact of distribution shifts on NN uncertainty estimates with applications in OR (i.e., shortcoming 2). 
That is, uncertainty estimates are always evaluated on benchmark datasets, e.g., MNIST and not-MNIST, or using artificial distortions, e.g., Gaussian blur \citep{ovadia2019can}.

\subsection{Uncertainty as explainable artificial intelligence}

\begin{figure}[h!]
	\centering
	\includegraphics[width=0.8\textwidth]{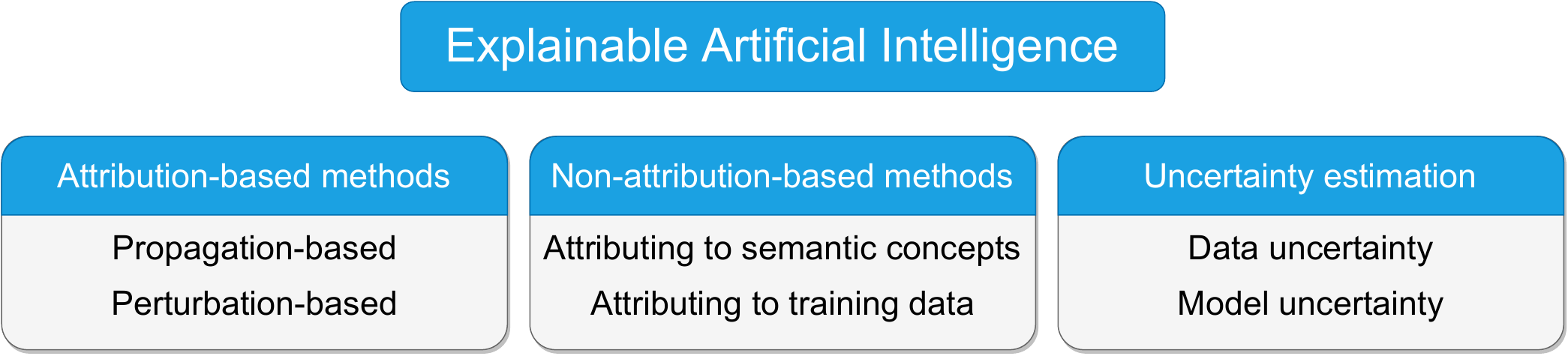}
	\caption{\textbf{Overview explainable artificial intelligence.} Uncertainty estimation is a third general type of XAI technique. Figure adapted from \citet{bai2021explainable}.}
	\label{fig:overview_xai}
\end{figure}

\citet{bai2021explainable} are the first to list uncertainty estimation as a third distinct category in XAI, next to attribution-based (e.g., SHAP) and non-attribution-based (e.g., counterfactual explanations) methods (Figure \ref{fig:overview_xai}). 
However, \citet{bai2021explainable} do not specify uncertainty estimation in terms of the two main XAI criteria and do not provide a theoretical motivation (i.e., shortcoming 3). 

This work addresses the three shortcomings by proposing a general uncertainty framework, positioning \textit{uncertainty estimation in ML models as XAI} and using \textit{classification with rejection}. 
Furthermore, a case study on \textit{NNs} with distribution shifts demonstrates the value of the framework for OR applications.

\section{Methodology}
\label{sec:methodology}
\begin{figure}[h!]
	\centering
	\includegraphics[width=0.8\textwidth]{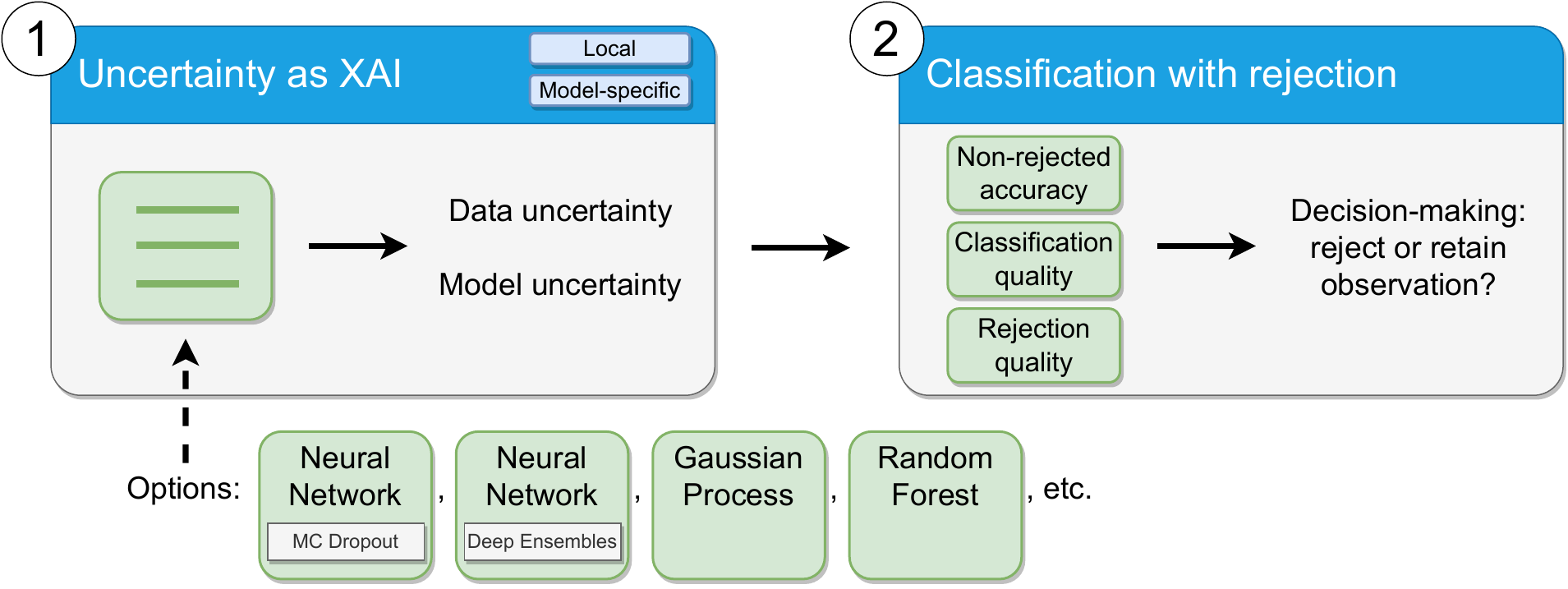}
	\caption{\textbf{General uncertainty framework.} The framework consists of two stages: (i) uncertainty estimation as XAI and (ii) classification with rejection. It can be applied to multiple ML models, each having one or more specific uncertainty techniques.}
	\label{fig:uncertainty_framework}
\end{figure}

Figure \ref{fig:uncertainty_framework} outlines the proposed general uncertainty framework.
The framework consists of two stages: (i) uncertainty estimation as XAI and (ii) classification with rejection. 
The goal of uncertainty estimation is to quantify the data and model uncertainty in predictions made by an ML model. 
The classification with rejection system then uses the estimates to assist in deciding which predictions should be rejected or retained, based on three key metrics.

\subsection{Uncertainty as explainable artificial intelligence}

Uncertainty as XAI is available for multiple ML models, each having distinct techniques.
That is, the case study quantifies data and model uncertainty in NNs, but it can also be computed in e.g., Gaussian Processes \citep{hullermeier2021aleatoric} or Random Forests \citep{shaker2020aleatoric} using other existing techniques.
Furthermore, one can even use different uncertainty estimation techniques for some ML models, e.g., Monte Carlo Dropout and Deep Ensembles for NNs \citep{gal2016dropout,lakshminarayanan2017simple}.

\subsubsection{Data and model uncertainty}

Each prediction has two uncertainty values, as uncertainty can arise from two fundamentally different sources: data uncertainty and model uncertainty \citep{der2009aleatory}.
\textit{Data uncertainty}, also known as aleatoric uncertainty, refers to the notion of randomness and is related to the data-measurement process. 
This uncertainty is irreducible even if more data is collected.
\textit{Model uncertainty}, also known as epistemic uncertainty, accounts for uncertainty in the model parameters, i.e., uncertainty about which model generated the collected data. 
In contrast to data uncertainty, collecting more data can reduce model uncertainty. 
Both types of uncertainty can then be summed to compute the \textit{total uncertainty} in a prediction.

\begin{figure}[h!]
	\centering
	\includegraphics[width=0.50\textwidth]{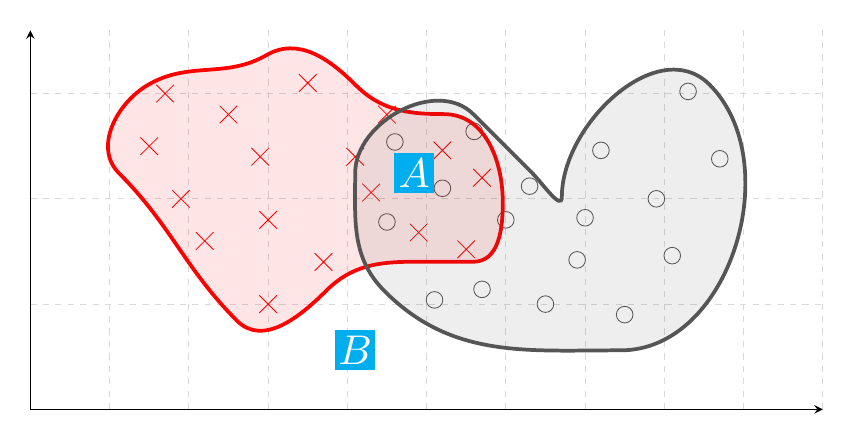}
	\caption{\textbf{Two types of uncertainty.} Observation \textit{A} has high data uncertainty; \textit{B} has high model uncertainty. Figure adapted from \citet{hullermeier2021aleatoric}.}
	\label{fig:types_uncertainty}
\end{figure}

Consider a binary classification task with two input features (Figure \ref{fig:types_uncertainty}), where the crosses represent positive training examples and the circles represent negative training examples.
At test time, predictions are made for both observations $A$ and $B$.
The model uncertainty is high in sparsely populated regions with few training examples. 
Therefore, observation \textit{A} has high model uncertainty and could be classified as either positive or negative. 
In contrast, observation \textit{B} lies in a region where the two class distributions are overlapping, i.e., the data uncertainty is high. 
Although collecting more training data around observation \textit{B} will reduce the model uncertainty, more training data around observation \textit{A} will not reduce the data uncertainty.

\subsubsection{Theoretical properties}

\citet{doshi2017towards} devise six desirable properties for XAI techniques: trust, actionability, fairness, privacy, robustness, and causality. 
We apply uncertainty estimation to this list and argue that it satisfies three properties:
\begin{itemize}
	\item Trust: decision makers should feel comfortable relinquishing control to the ML model. As model uncertainty enables saying ``I do not know,'' a human expert can step in. This awareness gives decision makers more confidence to rely on the model’s predictions in other situations when it says ``I do know.''
	\item Actionability: ML models should provide information assisting users to accomplish a task. Uncertainty estimates are key in classification with rejection, where uncertain observations are passed on to a human expert.
	\item Robustness: ML models should reach certain levels of performance in the face of input variation. Under increased distribution shift, uncertainty estimates grow accordingly, enabling an improvement in accuracy by rejecting the most uncertain observations.
\end{itemize}
To demonstrate the validity of the theoretical properties, they are evaluated in light of the case study results (see section \ref{sec:discussion}).
Uncertainty is complementary to other XAI techniques, which can be used to satisfy the remaining three properties.

\subsubsection{Specification}

Uncertainty estimation is a local and model-specific XAI method. 
It is \textit{local} because each observation receives an uncertainty estimate, both for data and model uncertainty. 
Furthermore, it is \textit{model-specific} because techniques for decomposition into data and model uncertainty are different across ML models, although they exist for multiple ML models.

\subsection{Classification with rejection}

For predictions with high uncertainty, the observations can be passed on to a human expert for a label.
The goal of a \textit{classification with rejection} system \citep{mena2021survey, barandas2022uncertainty} is to help decide when to stop rejecting the most uncertain observations.
The system takes as input the per-observation uncertainty values and outputs three metrics to assist the decision maker in finding the optimal rejection threshold for the task at hand.
It is useful for applications where making an error can be more costly than asking a human expert for help.
For example, in fraud detection, an employee can verify a transaction manually if the prediction is uncertain.

\begin{figure}[h!]
	\centering
	\includegraphics[width=0.60\textwidth]{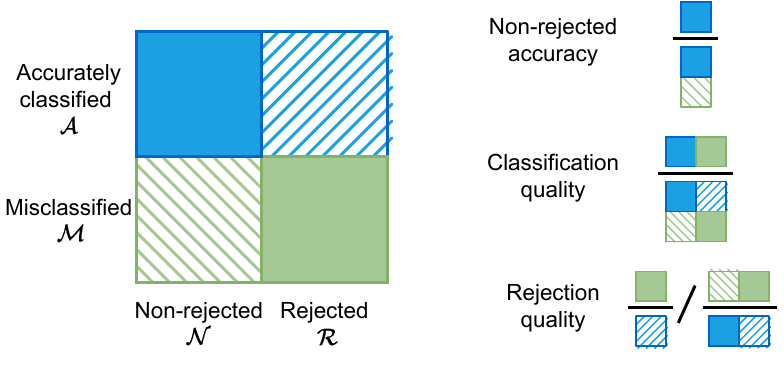}
	\caption{\textbf{Performance measures for classification with rejection.} Three performance measures are proposed by \citet{condessa2017performance} to find the optimal rejection point. Figure adapted from \citet{mena2021survey}.}
	\label{fig:rejection_confusion_matrix}
\end{figure}

Observations are classified along two criteria: (i) accurately classified $\mathcal{A}$ and misclassified $\mathcal{M}$; (ii) rejected $\mathcal{R}$ and non-rejected $\mathcal{N}$. 
\citet{condessa2017performance} propose three rejection metrics (Figure \ref{fig:rejection_confusion_matrix}; higher is better):
\begin{itemize}
	\item Non-rejected accuracy (NRA): ability to classify non-rejected samples accurately.
	\begin{equation}\label{NRA}
		NRA = \frac{|\mathcal{A} \cap \mathcal{N}|}{|\mathcal{N}|}
	\end{equation}
	\item Classification quality (CQ): ability to retain correctly classified samples and to reject misclassified samples, i.e., correct decision-making.
	\begin{equation}\label{CQ}
		CQ = \frac{|\mathcal{A} \cap \mathcal{N}| + |\mathcal{M} \cap \mathcal{R}|}{|\mathcal{N}| + |\mathcal{R}|}
	\end{equation}
	\item Rejection quality (RQ): ability to concentrate misclassified samples in the set of rejected samples.
	\begin{equation}\label{RQ}
		RQ = \frac{|\mathcal{M} \cap \mathcal{R}|}{|\mathcal{A} \cap \mathcal{R}|} / {\frac{|\mathcal{M}|}{|\mathcal{A}|}}
	\end{equation}
\end{itemize}

The NRA and CQ are bounded in the interval $[0,1]$, unlike the RQ which has a minimum value of zero and an unbounded maximum.
The three metrics are evaluated as a function of the percentage of rejections varying from 0\% to 100\% \citep{yong2022bayesian}.
If observations have identical uncertainty values (e.g., exactly \num{0.0} or \num{1.0}), observations are rejected randomly until the desired percentage is achieved.

\subsection{Workflow with a human-in-the-loop}

The suggested way of working for the human expert is as follows.
The accuracy score is first evaluated on the entire test set, without rejecting any observations.
If the accuracy is not sufficiently high, the rejection process is started and the metrics NRA, CQ, and RQ are evaluated.
The most uncertain observations are rejected until the labeling budget for the expert annotator is exhausted, or until the NRA is sufficiently high.
Although the NRA is the most important metric, the CQ and RQ provide more information on the internals of the rejection system.
For example, a decreasing CQ indicates that more and more correct observations are rejected.
At this point, the expert might decide to stop rejecting because the NRA will likely stagnate, which does not justify spending the labeling budget on.

It is important to note that the rejection level depends on the labeling budget available for the expert annotator and the accuracy requirement, associated with the misclassification cost, for the task. 
As such, there is no universally optimal point of rejection.

\section{Uncertainty in Neural Networks}
\label{sec:uncertainty_nn}
In the case study, NNs are used for the ``Uncertainty as XAI'' building block of the general uncertainty framework. 
This section discusses how uncertainty can be represented and measured in NNs.

\subsection{Data and model uncertainty}

\textbf{Data uncertainty.} In a NN classifier, the output layer contains a softmax or sigmoid function, forming a categorical distribution over the class labels $p(\mathbf{Y} \mid \mathbf{X},\ \bm{\theta})$. 
This distribution enables the NN to represent \textit{data uncertainty}.

Modern NNs are usually trained using a maximum likelihood objective. 
That is, they find a single setting of parameters $\bm{\theta}^*$ to maximize the probability of the data given the parameters, $\arg \max_{\bm{\theta}} p(\mathbf{X},\ \mathbf{Y} \mid \bm{\theta})$. 
For each test input $\mathbf{x}^*$, there is only one prediction because the NN generates an identical output for each run. 
As a result, model uncertainty cannot be captured.

\textbf{Model uncertainty.} NNs are large flexible models capable of representing many functions, corresponding to different parameter settings. 
Each function fits the training data well, yet generalizes in different ways, a phenomenon known as \textit{underspecification} \citep{wilson2020case}.
Considering all of these different NNs together allows capturing \textit{model uncertainty}. 
In a probabilistic sense, uncertainty in an unseen input point $\mathbf{x}^*$ is represented by the posterior predictive distribution $p(\mathbf{y}^* \mid \mathbf{x}^*,\ \mathbf{X},\ \mathbf{Y})$.

Model uncertainty can be captured in NNs using two approaches: (i) Bayesian NNs and (ii) ensembles. 
A Bayesian NN aims to estimate a full distribution for $p(\bm{\theta} \mid \mathbf{X},\ \mathbf{Y})$, unlike maximum likelihood.
However, this distribution is intractable and is typically approximated using sampling techniques. 
The ensembling approach obtains multiple good maximum likelihood settings $\bm{\theta}^*$. 

\begin{figure}[h!]
	\centering
	\includegraphics[width=0.60\textwidth]{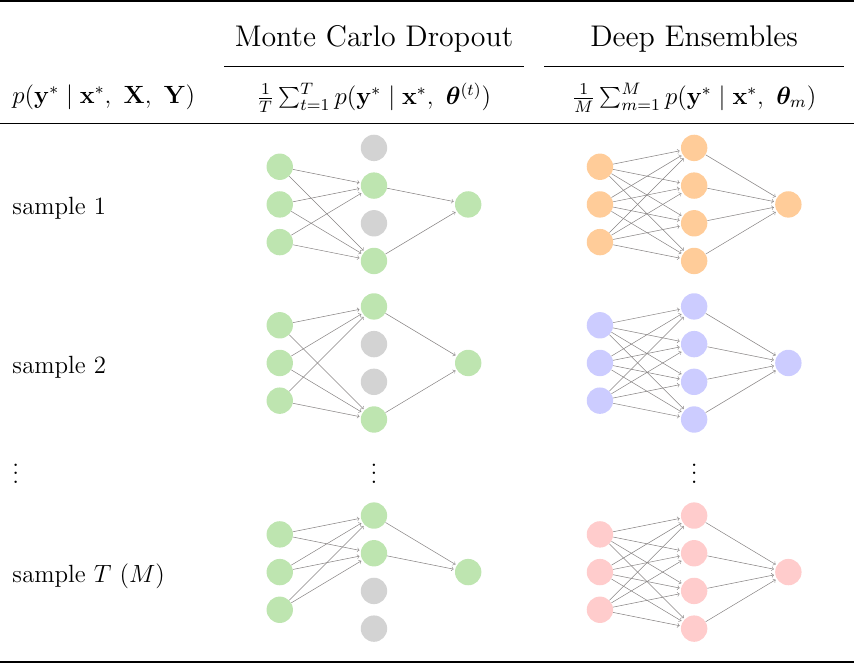} 
	\caption{\textbf{Overview of uncertainty estimation methods.} Forward passes are generated differently depending on the method. In MC dropout, different units are dropped out from a NN; in Deep Ensembles, multiple independent NNs are used, with different parameter initializations and noise in the SGD training process.}
	\label{fig:overview_uq_methods2}
\end{figure}

Both approaches aggregate predictions over a collection of NNs. 
The following subsections discuss the most popular technique for either approach, (i) Monte Carlo Dropout and (ii) Deep Ensembles.
Figure \ref{fig:overview_uq_methods2} provides a visual overview.

\subsection{Monte Carlo Dropout}

In Monte Carlo (MC) Dropout \citep{gal2016dropout}, dropout is not only applied at training time but also at test time. 
Multiple forward passes are performed, each time randomly dropping units and getting another thinned dropout variant of the NN. The resulting $T$ predictions $\{\widehat{\mathbf{y}}_1^*(\mathbf{x}^*),\ \dots ,\ \widehat{\mathbf{y}}_T^*(\mathbf{x}^*)\}$ are aggregated, forming an approximation to the true posterior predictive distribution:
\begin{align}\label{eq:mc_dropout}
	p(\mathbf{y}^* \mid \mathbf{x}^*,\ \mathbf{X},\ \mathbf{Y}) &= \int p(\mathbf{y}^* \mid \mathbf{x}^*,\ \bm{\theta}) p(\bm{\theta}\mid \mathbf{X},\ \mathbf{Y}) \,d\bm{\theta}\\
	&\approx \frac{1}{T} \sum_{t=1}^{T} p(\mathbf{y}^* \mid \mathbf{x}^*,\ \bm{\theta}^{(t)}).
\end{align}
The posterior predictive distribution is obtained through Bayesian model averaging. 
That is, it averages over an infinite collection of parameter settings, weighted by their posterior probabilities.

\subsection{Deep Ensembles}

Deep Ensembles \citep{lakshminarayanan2017simple} uses an ensemble of $M$ maximum likelihood NNs, with every NN trained on the same dataset and the same input features. 
The diversity arises through different parameter initializations and noise in the stochastic gradient descent (SGD) training process, inducing different solutions due to the non-convex loss. 
At test time, each of the $M$ NNs performs one forward pass. 
The resulting $M$ predictions $\{\widehat{\mathbf{y}}_1^*(\mathbf{x}^*),\ \dots ,\ \widehat{\mathbf{y}}_M^*(\mathbf{x}^*)\}$ are averaged, forming a mixture distribution:
\begin{equation}\label{eq:deep_ensembles}
	p(\mathbf{y}^* \mid \mathbf{x}^*) \approx \frac{1}{M} \sum_{m=1}^{M} p(\mathbf{y}^* \mid \mathbf{x}^*,\ \bm{\theta}_m).
\end{equation}

In ensembling, NNs are weighted equally over a finite collection of functions. 
As such, it is a fundamentally different mindset than Bayesian model averaging.

\subsection{Uncertainty decomposition}
\label{subsec:uncertainty_decomposition}

The posterior predictive distribution holds information about the total uncertainty in a prediction, decomposable in data and model uncertainty using classical information-theoretic measures. 
However, calculations require the expectation over the posterior distribution, which is intractable. 
Nonetheless, an approximation can be obtained using samples from the approximate posterior predictive distribution:
\begin{align}
	u_{total}(\mathbf{x}^*) &\approx H\left[\frac{1}{T} \sum_{t=1}^{T} p(\mathbf{y}^* \mid \mathbf{x}^*,\ \bm{\theta}^{(t)})\right]\\
	u_{data}(\mathbf{x}^*) &\approx \frac{1}{T} \sum_{t=1}^{T} H\left[p(\mathbf{y}^* \mid \mathbf{x}^*,\ \bm{\theta}^{(t)})\right]\\
	u_{model}(\mathbf{x}^*) &= u_{total}(\mathbf{x}^*) - u_{data}(\mathbf{x}^*).
\end{align}

First, total uncertainty and data uncertainty are calculated; then model uncertainty is obtained as the difference \citep{depeweg2018decomposition}. 
Total uncertainty is computed by averaging over the different samples and calculating the entropy $H$.
Data uncertainty is computed by calculating the entropy in each sample and averaging the entropies. 
This boils down to fixing a set of weights $\bm{\theta}^{(t)}$, i.e., considering a distribution $p(\mathbf{y}^* \mid \mathbf{x}^*,\ \bm{\theta}^{(t)})$, essentially removing the model uncertainty. 
Model uncertainty is high if the distribution $p(\mathbf{y}^* \mid \mathbf{x}^*,\ \bm{\theta}^{(t)})$ varies greatly for different weights $\bm{\theta}^{(t)}$.
Intuitively, data uncertainty measures uncertainty in the softmax classification on individual samples; model uncertainty measures how much the samples deviate \citep{hullermeier2021aleatoric, barandas2022uncertainty}.

\begin{table}[h!]
	\centering
	\caption{\textbf{Examples of uncertainty decomposition.} The middle and bottom row have equal total uncertainty but have wildly different samples. Decomposition in data and model uncertainty reveals the different characteristics.} 
	\label{tab:decomp}
	\begin{tabular}{L{7.1cm} C{1.7cm} C{1.6cm} C{1.6cm} C{1.8cm}}\toprule
		Samples $p(\mathbf{y}^* \mid \mathbf{x}^*,\ \bm{\theta}^{(t)})$ & $p(\mathbf{y}^* \mid \mathbf{x}^*)$ & $u_{total}(\mathbf{x}^*)$ & $u_{data}(\mathbf{x}^*)$ & $u_{model}(\mathbf{x}^*)$\\
		\midrule
		\{(1.\phantom{0}, 0.\phantom{0}), (1.\phantom{0}, 0.\phantom{0}), (1.\phantom{0}, 0.\phantom{0}), (1.\phantom{0}, 0.\phantom{0})\} & (1.\phantom{0}, 0.\phantom{0}) & 0. & 0. & 0.\\
		\{(0.5, 0.5), (0.5, 0.5), (0.5, 0.5), (0.5, 0.5)\} & (0.5, 0.5) & 1. & 1. & 0.\\
		\{(1.\phantom{0}, 0.\phantom{0}), (0.\phantom{0}, 1.\phantom{0}), (1.\phantom{0}, 0.\phantom{0}), (0.\phantom{0}, 1.\phantom{0})\} & (0.5, 0.5) & 1. & 0. & 1.\\
		\bottomrule
	\end{tabular}
\end{table}

Table \ref{tab:decomp} contains three examples in the context of binary classification with $T=4$ samples. 
The middle and bottom rows both have a total uncertainty of \num{1.0} although the samples are wildly different. 
Therefore, the total uncertainty alone is not sufficient to characterize the NN's predictions; decomposition into data and model uncertainty is necessary.

\section{Case Study: Student Performance Prediction}
\label{sec:case_study}
\subsection{Problem setting}

Student performance prediction is extensively discussed in OR literature \citep{delen2020development, coussement2020predicting, olaya2020uplift, deeva2022predicting, phan2023decision}. 
Common performance metrics include student dropout, course certification, final course grade, pass/fail, etc. 
Developing predictive models for student performance forms the basis for an educational early-warning system, where at-risk students are identified on time and assisted with personalized support by course advisors. 
Therefore, in order to deliver support, a predictive model should provide predictions being both \textit{accurate} and \textit{actionable}. 

\citet{whitehill2017mooc} and \citet{gardner2018student} argue that most prior research has poor actionability due to same course--same year evaluation.
This training paradigm creates a practical problem because the target labels required by supervised learning algorithms only become available after the final exam, when any support for students is too late. 
Alternatives that resolve this issue are training on a previous edition of the course, or training on a different course altogether if there is no previous edition available. 

The case study applies the uncertainty framework to NNs and investigates uncertainty estimation as an XAI technique in a predictive setup subject to distribution shifts.
The results are compared to a standard NN only capable of capturing data uncertainty, but no model uncertainty.
The experiment answers the call of \citet{gavsevic2016learning} for research on changing course conditions in student performance prediction, advocating that learning analytics should account for the fluid nature of technology use within a course.

\subsection{Data}

The dataset \citep{DVN/26147/OCLJIV_2014} consists of student-course records of HarvardX and MITx massive open online courses (MOOCs) hosted on the edX platform about a wide range of topics, over two semesters (fall 2012 and spring 2013). 
The binary target labels denote whether a student scored a grade high enough to earn a certificate; features include processed clickstream activities and student demographics. 

\subsection{Experimental setup}

\begin{figure}[t!]
	\centering
	\includegraphics[width=0.8\textwidth]{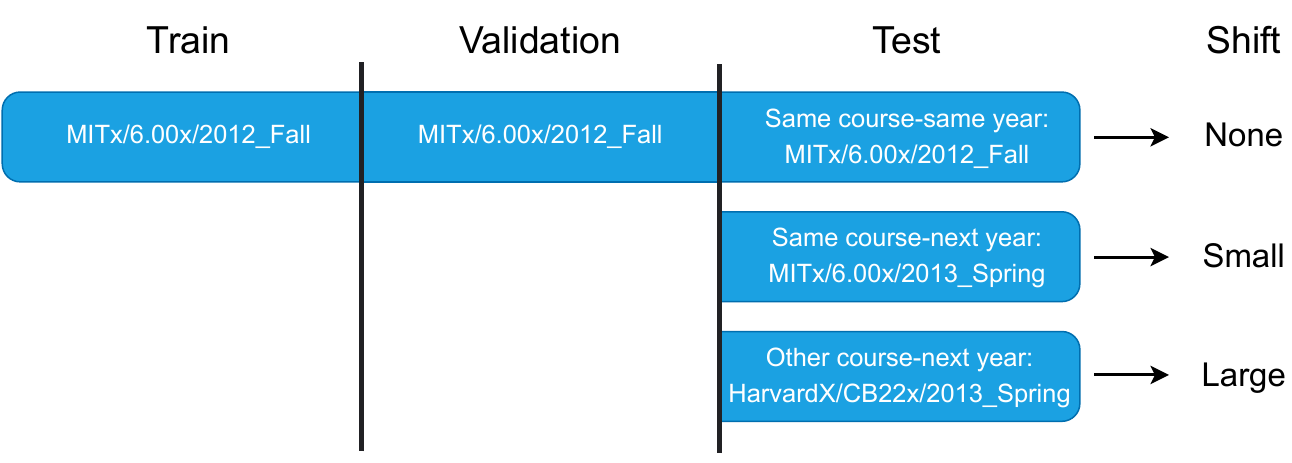}
	\caption{\textbf{Experimental setup.} A NN is trained on the course ``MITx/6.00x/2012\_Fall''. Next, predictions are made for three test sets: (i) ``MITx/6.00x/2012\_Fall'' (same course--same year), (ii) ``MITx/6.00x/2013\_Spring'' (same course--next year), (iii) ``HarvardX/CB22x/2013\_Spring'' (other course--next year).}
	\label{fig:experimental_setup}
\end{figure}

The experimental setup is detailed in Figure \ref{fig:experimental_setup}. 
First, the NN is trained on the course ``MITx/6.00x/2012\_Fall'', denoting the MITx course 6.00x ``Introduction to Computer Science and Programming'' of fall 2012. 
This course is selected because it has the largest number of observations and is offered in both semesters. 
Next, predictions are made on three test sets: (i) same course--same year: ``MITx/6.00x/2012\_Fall'', (ii) same course--next year: ``MITx/6.00x/2013\_Spring'', and (iii) other course--next year: ``HarvardX/CB22x/2013\_Spring''. 
Course CB22x is titled ``The Ancient Greek Hero'' and is selected as an extreme case because it is a non-STEM course offered by a different university. 
It is important to note that the input features gathered for HarvardX and MITx courses are identical because they are both hosted on the edX platform.

Predicting on the three test sets represents three different distribution shifts. 
This ranges from (i) no shift (same course--same year), to (ii) small shift (same course--next year), to (iii) large shift (other course--next year).
The situation of no shift serves as a baseline because it is often used in literature, despite being practically infeasible.

The data include students who accessed at least half of the chapters in the course material, with the training set having a class distribution of 56/44 and \num{2000} observations. 
The three test sets have a class distribution of (i) 56/44, (ii) 46/54, and (iii) 72/28, respectively. 

All three methods (standard, MC Dropout, Deep Ensembles) use the same NN configuration as the building block. 
The NN is a multi-layer perceptron with 2 hidden layers, each containing 64 hidden units, a ReLU activation function, a dropout rate of \num{0.4} and \num{0.5}, and Glorot uniform weight initialization. 
The NNs are trained with the Adam optimizer and a binary cross-entropy loss function for 50 epochs with a batch size of 32 and a learning rate of \num{5e-4}, using early stopping. 
For MC Dropout, the NN predicts 128 samples per input observation. 
For Deep Ensembles, 10 NNs are trained, resulting in 10 samples per input observation.
Results are averaged over 10 runs with random data splits.

\section{Results}
\label{sec:results}
\subsection{Total uncertainty}

\begin{figure}[t!]
	\centering
	\includegraphics[width=0.90\textwidth]{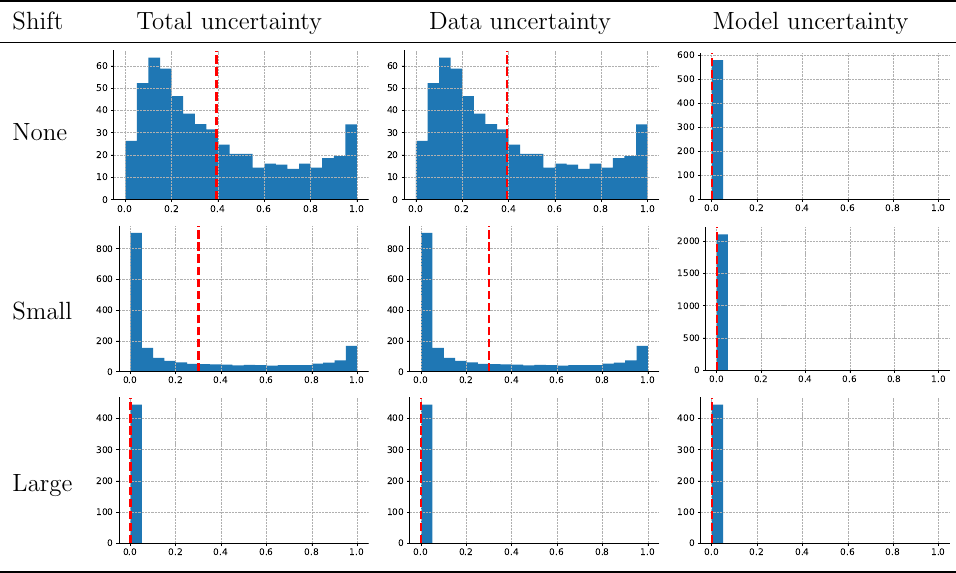} 
	\caption{\textbf{Standard: uncertainty distributions.} The histogram displays absolute frequency and the dashed line denotes the mean value. With increased distribution shift, data uncertainty decreases so total uncertainty decreases as well because the standard NN does not capture model uncertainty.}
	\label{fig:unc_distr_std}
\end{figure}

\begin{figure}[t!]
	\centering
	\includegraphics[width=0.90\textwidth]{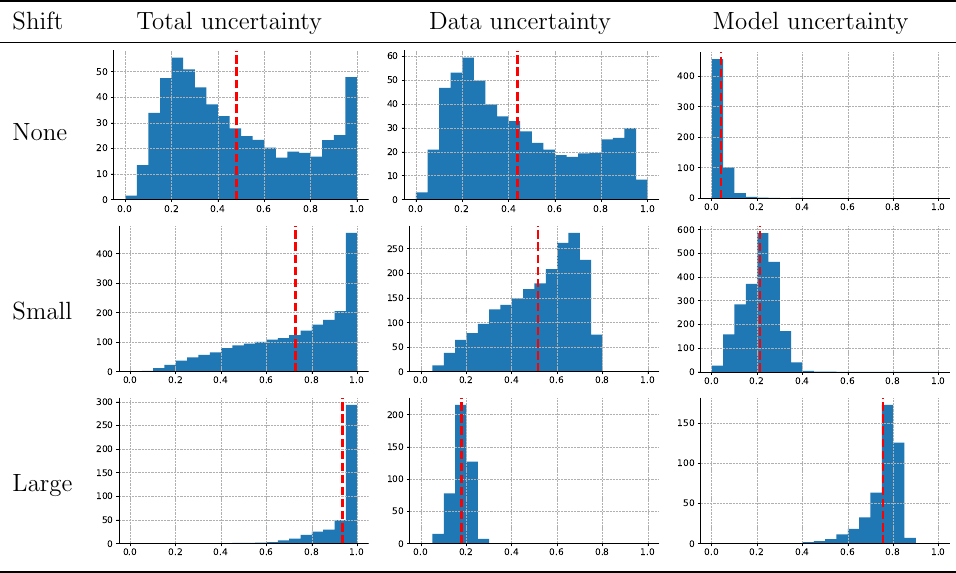} 
	\caption{\textbf{MC Dropout: uncertainty distributions.} The histogram displays absolute frequency and the dashed line denotes the mean value. With increased distribution shift, data uncertainty stagnates or decreases while model uncertainty increases consistently. As a result, MC Dropout has increased total uncertainty.}
	\label{fig:unc_distr_mcd}
\end{figure}

\begin{figure}[t!]
	\centering
	\includegraphics[width=0.90\textwidth]{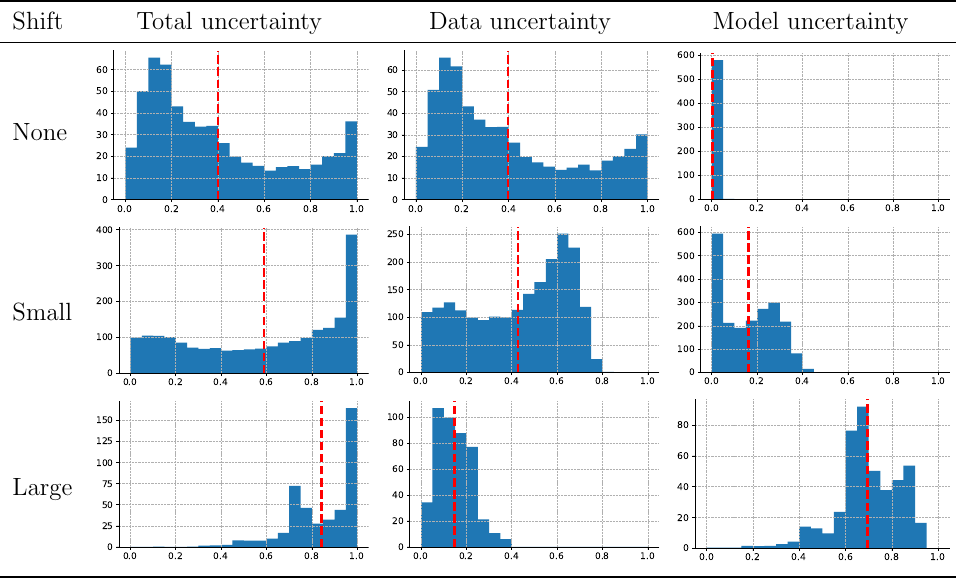} 
	\caption{\textbf{Deep Ensembles: uncertainty distributions.} The histogram displays absolute frequency and the dashed line denotes the mean value. With increased distribution shift, data uncertainty stagnates or decreases while model uncertainty increases consistently. As a result, Deep Ensembles has increased total uncertainty.}
	\label{fig:unc_distr_de}
\end{figure}

Figures \ref{fig:unc_distr_std}, \ref{fig:unc_distr_mcd}, and \ref{fig:unc_distr_de} show the histograms of the uncertainty distributions for all three methods on all three shifts (rows); the total uncertainty is decomposed into data and model uncertainty (columns). 
The vertical dashed line denotes the mean of the uncertainty values, which can be used to quickly compare centrality across distributions.

With increased distribution shift, data uncertainty consistently decreases for the standard NN, i.e., more mass is located at low uncertainty values and the mean value decreases.
For MC Dropout and Deep Ensembles, data uncertainty remains equal when moving to the small shift before decreasing substantially on the large shift.
In contrast to data uncertainty, model uncertainty grows rapidly for MC Dropout and Deep Ensembles. 
The standard NN cannot capture model uncertainty (i.e., value zero for all observations) and only relies on the decreasing data uncertainty to calculate total uncertainty. 

In summary, for increased distribution shift, the standard NN has decreased total uncertainty, whereas MC Dropout and Deep Ensembles have rapidly increased total uncertainty.
In other words, the standard NN becomes more confident as the inputs stray away from the training data distribution, which is undesirable behavior. 
This is in contrast to MC Dropout and Deep Ensembles, which indicate that the NN ``knows what it does not know.''

\begin{table}[h!]
	\centering
	\caption{\textbf{Accuracy (\%).} Accuracy degrades with increased distribution shift for all NNs. For a small and large shift, MC Dropout and Deep Ensembles outperform the standard NN. Mean $\pm$ standard error are reported.}
	\label{tab:accuracy}
	\begin{tabular}{L{1.5cm} C{2.5cm} C{2.5cm} C{3.0cm}}\toprule
		\addlinespace
		Shift & Standard & MC Dropout & Deep Ensembles \\
		\midrule
		\addlinespace
		None & $88.17 \pm 0.10$ & $87.81 \pm 0.25$ & $88.53 \pm 0.11$ \\
		\addlinespace
		Small & $63.85 \pm 0.92$ & $84.43 \pm 1.04$ & $84.93 \pm 0.52$ \\
		\addlinespace
		Large & $27.93 \pm 0.00$ & $45.05 \pm 4.30$ & $31.01 \pm 1.57$ \\
		\bottomrule
	\end{tabular}
\end{table}

\subsection{Accuracy}
\label{subsec:accuracy}

Table \ref{tab:accuracy} shows the accuracy for all three methods (columns) on all three shifts (rows). It is important to note that MC Dropout and Deep Ensembles average over the different samples to get the final probability vector, capturing model uncertainty. 

On increasingly shifted data, the accuracy degrades for all three methods, as expected. 
For no shift, MC Dropout has a slightly lower accuracy than the standard NN while Deep Ensembles performs slightly better.
For a large shift, all NNs perform poorly because they tend to naively predict the minority class, which illustrates how difficult the task is.
The results of MC Dropout also have a larger standard error in this situation, indicating that the model results alternate between predicting the minority class and making more sensible predictions.
In the situation of a small shift, information contained in different samples (i.e., model uncertainty) has a big impact on accuracy; MC Dropout and Deep Ensembles improve substantially over the standard NN. 
That is, the standard NN has an accuracy of 63.85\%, MC Dropout has 84.43\%, and Deep Ensembles has 84.93\%.

It is worth noting that although all methods have poor accuracy under a large shift, MC Dropout and Deep Ensembles raise a warning through increased uncertainty whereas the standard NN is confidently wrong. 
The uncertainty values are then used to reject the most uncertain observation, i.e., classification with rejection.

\subsection{Non-rejected accuracy}

\begin{figure}[tb!]
	\centering
	\includegraphics[width=0.90\textwidth]{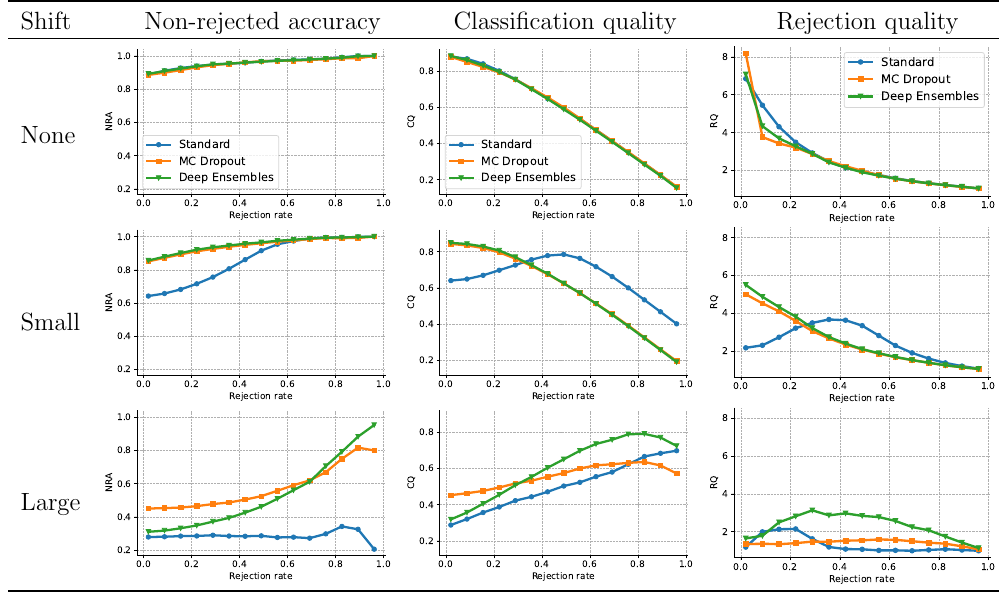}
	\caption{\textbf{Classification with rejection.} NRA, CQ, and RQ are displayed for increased distribution shifts, for all three models. Under a large distribution shift, the standard NN's uncertainty estimates are uninformative and rejections are random. This is evidenced by the stagnating NRA, ever-increasing CQ and low RQ.}
	\label{fig:3_metrics_plot}
\end{figure}

Figure \ref{fig:3_metrics_plot} (left column) displays the non-rejected accuracy for all three methods on all three shifts (rows), based on the total uncertainty. 
Note that the curve at rejection \num{0.0}\% corresponds to the method's accuracy without rejection in Table \ref{tab:accuracy}.

In the situation of no shift (top row), all three methods achieve 100\% accuracy.
The same holds for a small shift (middle row), despite that the standard NN started at a substantially lower initial accuracy. 
For a large shift (bottom row), initial accuracies are all poor but only MC Dropout and Deep Ensembles manage to increase the accuracy by rejecting the most uncertain observations. 
Deep Ensembles performs better with 95\% accuracy at rejection rate 0.95, whereas MC Dropout only obtains 80\% accuracy.
The standard NN, in contrast, has a large amount of observations with uncertainty zero. 
Since these uncertainty values are identical, observations are rejected randomly, causing the non-rejected accuracy to stagnate at the initial accuracy.

Only MC Dropout and Deep Ensembles have informative uncertainty values so that appropriate observations are rejected, effectively increasing the NRA. 
The standard NN, on the other hand, fails to increase the NRA under large distribution shifts.

\subsection{Classification quality}

Figure \ref{fig:3_metrics_plot} (middle column) shows the classification quality for all three methods on all three shifts (rows), based on the total uncertainty. 
Classification quality measures the correct decision-making of the classifier-rejector; accurately classified samples should be retained and misclassified samples should be rejected. 
In other words, the curve shows where the maximum number of correct decisions is made.

In the situation of no shift (top row), 
all three curves decrease gradually, indicating that the majority of rejected observations are correctly classified. 
As such, it is optimal to not reject any observations.
When a distribution shift is present (middle and bottom row), MC Dropout and Deep Ensembles obtain the point of optimal decision-making at smaller rejection rates than the standard NN. 
For the small shift (middle row), MC Dropout and Deep Ensembles obtain the maximum CQ at rejection rate 0\%; the standard NN needs 50\% rejections.

For the large shift (bottom row), the CQ curve of the standard NN keeps on increasing, i.e., it would be best to reject all observations. 
In contrast, MC Dropout and Deep Ensembles obtain the maximum value at rejection rate 80\%.
At this rate, Deep Ensembles outperforms MC Dropout with a CQ of 0.80, as compared to MC Dropout's CQ of 0.63.
These findings indicate that the uncertainty estimates of the standard NN are least effective at deciding whether observations should be retained or rejected, and that Deep Ensembles is more effective than MC Dropout.

\subsection{Rejection quality}

Figure \ref{fig:3_metrics_plot} (right column) displays the rejection quality for all three methods on all three shifts (rows), based on the total uncertainty.
Rejection quality measures the ability to reject misclassified samples. 
That is, it compares the proportion of misclassified to accurately classified samples on the set of rejected samples with that proportion on the entire data set.

In the situation of no shift (top row), 
all three curves decrease rapidly, indicating that the proportion of misclassified observations in the rejected set decreases as more observations are rejected, which is undesirable.
For the small shift (middle row), MC Dropout and Deep Ensembles obtain the highest RQ at rejection rate 0\%. 
In contrast, the standard NN requires 40\% rejections to do the same.

For the large distribution shift (bottom row), Deep Ensembles outperforms the other two methods by achieving much higher RQ values.
This finding indicates that the majority of rejected observations are misclassified, effectively improving the NRA. 
MC Dropout has lower RQ values, with the NRA curve increasing slower.
The standard NN falls quickly to RQ value \num{1.0}, indicating that the uncertainty estimates are uninformative and rejections are random. 
This trend is reflected in the stagnating NRA.

\section{Discussion}
\label{sec:discussion}
Methods with uncertainty estimation as XAI perform on par or better than the standard NN.
If there is no shift, there is no difference between the methods. 
However, from the point a distribution shift is present, uncertainty estimation leads to optimal decision-making at smaller rejection rates. 
For small shifts, capturing model uncertainty induces higher initial accuracy and fewer rejections to obtain a specific level of accuracy. 
For large distribution shifts, it issues a warning about novel observations so the system can reject predictions accordingly, unlike the standard NN.

Uncertainty as XAI increases \textit{trust} in an ML system by also capturing model uncertainty, indicating when an observation lies outside the observed training data. 
The results show that the model uncertainty values are sensitive to changes in the data distribution, providing an important source of information to the decision maker not available in the standard NN.
Furthermore, \textit{robustness} is improved as the total uncertainty grows with increasing distribution shifts, while uncertainty values in the standard NN decrease. 
Finally, \textit{actionability} is increased by directly using the uncertainty information in a classification with rejection system, raising the NRA even under large distribution shifts.

Continuing on the increased actionability, the decision maker should inspect Figure \ref{fig:3_metrics_plot} to decide on the appropriate rejection threshold given the specific labeling budget and accuracy requirements.
For example in the situation of no shift, it would be sensible to label at most 20\% of the observations as the CQ and RQ decrease quickly, resulting in a slowly increasing NRA curve.
In contrast, for the large shift, the full labeling budget can be used as the NRA continues to improve.

\section{Conclusion}
\label{sec:conclusion}
This paper proposes a general uncertainty framework positioning \textit{uncertainty estimation in ML models as an XAI technique}, giving local and model-specific explanations.
Furthermore, the framework uses \textit{classification with rejection} to reduce misclassifications by bringing a human expert in the loop for uncertain observations.
Finally, the framework is applied to a case study of \textit{NNs} in educational data mining subject to distribution shifts.

The case study demonstrates that standard NNs only capturing data uncertainty are confidently wrong when confronted with distribution shifts.
In contrast, NNs equipped with uncertainty estimation as XAI raise a warning in novel situations through increased model uncertainty, offering a solution to their overconfidence. 
Deep Ensembles outperform MC Dropout as an XAI technique with higher quality uncertainty estimates, obtaining higher accuracy when rejecting the most uncertain observations.
Uncertainty as XAI improves the model's \textit{trustworthiness} in downstream decision-making tasks, giving rise to more \textit{actionable} and \textit{robust} ML systems in OR.

Several directions for future work are possible. 
The case study only considers knowing the target labels in time due to limitations in the data; studies also satisfying the requirement for input features would help validate the findings.
Although this paper focuses on uncertainty for NN classifiers, uncertainty can also be quantified for NN regression models and other ML models such as Gaussian Processes \citep{price2019gaussian} and Random Forests \citep{shaker2020aleatoric}.
Finally, uncertainty as XAI can be used in active learning, where limited labeled training data is available and the ML system can ask a human expert to label the most uncertain observations \citep{kadzinski2021active}.

\section*{Acknowledgments}
This work was supported by the Research Foundation Flanders (FWO) [grant number 1S97022N].

% =====================================================================
% End matter
% =====================================================================

%\bibliographystyle{unsrtnat}
%\bibliography{references.bib}  %%% Uncomment this line and comment out the ``thebibliography'' section below to use the external .bib file (using bibtex) .

% Uncomment this section and comment out the \bibliography{references} line above to use inline references.

\end{document}